\documentclass[PHM, 2023]{PHMSociety}

\usepackage{graphicx}
\usepackage{amsmath}
\usepackage{xcolor}
\usepackage{multirow}

\begin{document}

\title{A Closer Look at Bearing Fault Classification Approaches}

\author{%
	Harika Abburi\authorNumber{1} , Tanya Chaudhary\authorNumber{1}, Haider Ilyas\authorNumber{2}, Lakshmi Manne\authorNumber{2}, Deepak Mittal \authorNumber{1}, \\ Don Williams \authorNumber{3}, Derek Snaidauf \authorNumber{3}, Edward Bowen\authorNumber{2}, Balaji Veeramani\authorNumber{2}{*}
}
\address{
	\affiliation{{1}}{
Deloitte \& Touche Assurance \& Enterprise Risk Services India Private Limited, India}{ 
		{\email{abharika@deloitte.com}}, {\email{tanychaudhary@deloitte.com}}, {\email{deepamittal@deloitte.com}}
		} 
	\tabularnewline 
	\affiliation{2}{Deloitte \& Touche LLP, USA}{ 
		{\email{hailyas@deloitte.com}}, {\email{lamanne@deloitte.com}}, {\email{edbowen@deloitte.com}}, {\email{bveeramani@deloitte.com}}
		}
	\tabularnewline 
	\affiliation{3}{Deloitte Transactions \& Business Analytics LLP, USA}{ 
		{\email{dowilliams@deloitte.com}},{\email{dsnaidauf@deloitte.com}}
		}
		\affiliation{}{}{ 
		{\email{*\textit{Corresponding author}}}
		}

	}

\maketitle



\begin{abstract}
Rolling bearing fault diagnosis has garnered increased attention in recent years owing to its presence in rotating machinery across various industries, and an ever increasing demand for efficient operations. Prompt detection and accurate prediction of bearing failures can help reduce the likelihood of unexpected machine downtime and enhance maintenance schedules, averting lost productivity.  Recent technological advances have enabled monitoring the health of these assets at scale using a variety of sensors, and predicting the failures using modern Machine Learning (ML) approaches including deep learning architectures. Vibration data has been collected using accelerated run-to-failure of overloaded bearings, or by introducing known failure in bearings, under a variety of operating conditions such as rotating speed, load on the bearing, type of bearing fault, and data acquisition frequency. However, in the development of bearing failure classification models using vibration data there is a lack of consensus in the metrics used to evaluate the models, data partitions used  to evaluate models, and methods used to generate failure labels in run-to-failure experiments. An understanding of the impact of these choices is important to reliably develop models, and deploy them in practical settings. In this work, we demonstrate the significance of these choices on the performance of the models using publicly-available vibration datasets, and suggest model development considerations for real world scenarios. Our experimental findings demonstrate that assigning vibration data from a given bearing across training and evaluation splits leads to over-optimistic performance estimates, PCA-based approach is able to robustly generate labels for failure classification in run-to-failure experiments, and $F$ scores are more insightful to evaluate the models with unbalanced real-world failure data.




 
\end{abstract}

\section{Introduction}

Bearings are a crucial component of machines used across various industries, and their reliable operation is critical for an organization to robustly maintain its  supply chain. Unplanned downtime of machines leads to  revenue loss, lowered productivity, and missed production targets which in turn adversely affects an organization’s ability to meet its obligations leading to a reputation and other risks. However, bearings are prone to failure for a variety of reasons, including material defects, corrosion, wear, and poor installation \shortcite{howard1994review}. Predicting bearing failures in advance, and replacing the bearing to avoid inflicting significant damage to the machine will help lower both maintenance and capital costs. Vibration signals obtained using a variety of accelerometers are widely used to monitor and assess the health of rotatory systems \shortcite{murphy2020choosing}. The scale at which  machines are deployed across various industries requires automated ways to monitor failures arising in these machines and alert maintenance personnel \shortcite{fausing2020predictive}.   

Automated detection of bearing failures \cite{howard1994review} and the estimation of a component's remaining usable life \shortcite{schwendemann2021survey} have been active research topics for many years, however the technologies to monitor and detect failures at scale has become possible by advances in internet of things (IoT) and artificial intelligence (AI) \shortcite{zhao2020deep}. Monitoring the health of thousands of machines has become possible, with a wide variety of vibration sensors (both piezoelectric and Micro Electro Mechanical Systems (MEMS) accelerometers) using a variety of signal acquisition characteristics, battery life, and data communication capabilities to on-premise or cloud servers. Automated detection approaches have evolved from basic statistical approaches to more modern learning based approaches. Statistical methods in the time domain use energy level (root mean square value, crest factor), kurtosis, peak, and shape of the amplitude probability distribution, to detect failures \shortcite{yazdi2019experimental}. Spectral features provide a complementary view of the vibration waveform that readily shows differences between various bearing failures \cite{xu2020imbalanced}. Time-frequency approaches that are effective in non-stationary signals,such as the short-time Fourier transform (STFT) and Wavelet Transform (WT) have also been used to analyze vibration signals \shortcite{wang2017virtualization}. In addition, other orthogonal transforms such as Continuous Wavelet Transform (CWT), Discrete Wavelet Transform (DWT), and Empirical Mode Decomposition (EMD) \shortcite{buchaiah2022bearing} have been explored. Condition monitoring approaches used heuristics/thresholds (e.g. ISO 20816) on these features to detect and classify failure, however the performance of these approaches is limited as these thresholds don't necessarily generalize to different operating settings. Machine learning (ML) approaches have helped overcome these limitations by learning to discriminate the failures automatically from data.

ML approaches have used features extracted from vibration signals along with support vector machine (SVM, Random Forest (RF), XGBoost, and various other classifiers \shortcite{tyagi2008comparative}, and demonstrated improved performance in detecting various faults. Recently, Deep Learning (DL) approaches, which have demonstrated superior performance across several large-scale benchmarks in computer vision and natural language processing \shortcite{lecun2015deep}, have garnered increased attention in the Prognostics and Health Monitoring (PHM) community for detecting failures of rotating machinery.  A variety of DL architectures including Convolutional Neural Networks (CNN) and its variants, Recurrent Neural Networks (RNN), Auto Encoder (AE), Generative Adversarial Networks (GANs) have been used for detecting failures. However there is a lack of consensus on choice of how the machine learning problem is formulated, model is trained, or evaluated across these studies. 

Bearing fault classification problems are formulated as binary or multi-class classification problems with outputs as bearing labels \shortcite{zhao2020deep}, failure/no-failures \shortcite{cui2022feature}, or different failure classes \shortcite{zhao2021improved}. Several studies that were developed to detect failures have leveraged popular open source datasets \shortcite{wang2018hybrid, berghout2021semi, hendriks2022towards}. Vibration data available in these datasets have been gathered with experimental setups, using accelerated run-to-failure of overloaded bearings or manually damaged bearings, under a variety of operating conditions. In order to accurately assess performance of these bearing classification models, input data has to be split into training, validation and test partitions without any  information leakage between the splits \shortcite{abu2012learning}. Information leakage leads to over-optimistic performance estimates of models whose performance fails to hold in real-world scenarios \shortcite{abu2012learning, riley2019three}. Several fault classification studies \shortcite{zhao2020fault, ruan2023cnn} that have reported high performance however assign waveform recordings from the same bearing to both training and test partitions. We demonstrate such assignment of a bearing data across partitions leads to high performance estimates. In the run-to-failure experiments, the rate of degradation of the bearings is variable. Lesser number of data samples are collected from bearings that fail faster (assuming a fixed vibration data acquisition rate) which in turn results in lesser amount of data available to train the failure detection models. Further the criteria used by various studies to segment the run-to-failure data into failure and normal operating region, such as considering only last few samples in the wave file \shortcite{zhao2020deep}, or using a Principal Component Analysis (PCA) approach \shortcite{juodelyte2022predicting} adds to the variation and imbalance in the data available across classes. The number of bearings with particular injected failures further influences the amount of data available to train fine grained failure detection models. In spite of these differences in the number of bearings associated with a failure, and the amount of data available in the failure region of a bearing, accuracy has been widely used as performance metric by theses studies \shortcite{schwendemann2021survey,neupane2020bearing}, a choice not ideal in evaluating classifiers with unbalanced datasets and multi-class classifiers \shortcite{davis2006relationship}. Our work considers the effect of these choices on the failure prediction models.

In this work, we take a closer look at the formulation of the ML problem, the dataset apportioning choice for model development, and the metrics to evaluate failure classification models. We formulate the failure classification problem as a coarse failure/no-failure binary classification, or as more fine grained failures/no-failure multi-class classification to study their efficacy on different datasets. We investigate this using three bearing failure datasets (both run-to-failure and injected failures). We also demonstrate how the choice of segmenting the run-to-failure datasets using a threshold or an unsupervised PCA followed by a k-means clustering method influences the amount of data available for training the model.  We further investigate the influence of training, validation, and test dataset splits that considers the bearing information, and importance of using metrics (precision, recall, F-score, and $F_{mac}$)  in addition to accuracy. This work helps underscore the importance of several key choices in reliably developing models, and deploying them in practical settings for PHM.

The rest of this paper is organized as follows. In Section~\ref{sec:datasets} we briefly describe the datasets that are explored in this study followed by the proposed approach in Section~\ref{sec:approach}. In Section~\ref{results}, we demonstrate through experiments the influence of various choices, and finally summarize the effect of these choices in Section~\ref{sec:conclusion}.

\section{Dataset}
\label{sec:datasets}
This section describes three publicly-available bearing fault vibration datasets that we use in our study. 

\subsection{FEMTO}
The FEMTO dataset \shortcite{berghout2021semi} contains vibration data recorded during accelerated bearing failures using the PRONOSTIA experimental platform. Vibration data was collected in horizontal and vertical directions  using 17 bearings under three distinct operating conditions. In our study, we only consider horizontal vibration data where changes related to failures were observed. The vibration signal was recorded every ten seconds at a sampling rate of 25.6 kHz for 0.1 seconds resulting in 2560 data points per waveform. 

\subsection{XJTU}
Xi’an Jiaotong University dataset (XJTU) \shortcite{wang2018hybrid}  is a more recent vibration dataset collected during accelerated bearing failures. This dataset consists of data obtained from 15 run-to-failure bearings under three distinct operating conditions, where different bearing failure types have been identified. We use the failure types Inner race (IR), outer race (OR), Inner race-outer race (IR  and OR), cage, Inner race-ball-cage-outer race (COBI) failures along with normal data for our multi-class classification formulation. The vibration signal was recorded every minute at a sampling rate of 25.6 kHz for 1.28 seconds resulting in 32768 data points per waveform.   

\subsection{CWRU}
Case Western Reserve University (CWRU) dataset \shortcite{hendriks2022towards} contains vibration data collected from bearings with induced faults from the fan-end or the drive end bearings at different operating conditions. Single-point faults were introduced with fault diameters of 0.007, 0.014, and 0.021 inches on the rolling element (BALL), IR, and OR, respectively.  We use these failure categories along with normal operation as labels for the multi-class classification formulation. In this study a single waveform was recorded at either 12KHz or 48 KHz continuously for roughly 5-10 seconds.  We downsampled the 48KHz waveforms to 12 KHz in order to analyze data consistently regardless of the sampling frequency used in the recording.     

\section{Approach}
\label{sec:approach}

\begin{figure}[h]
\centering
\includegraphics[width=\columnwidth]{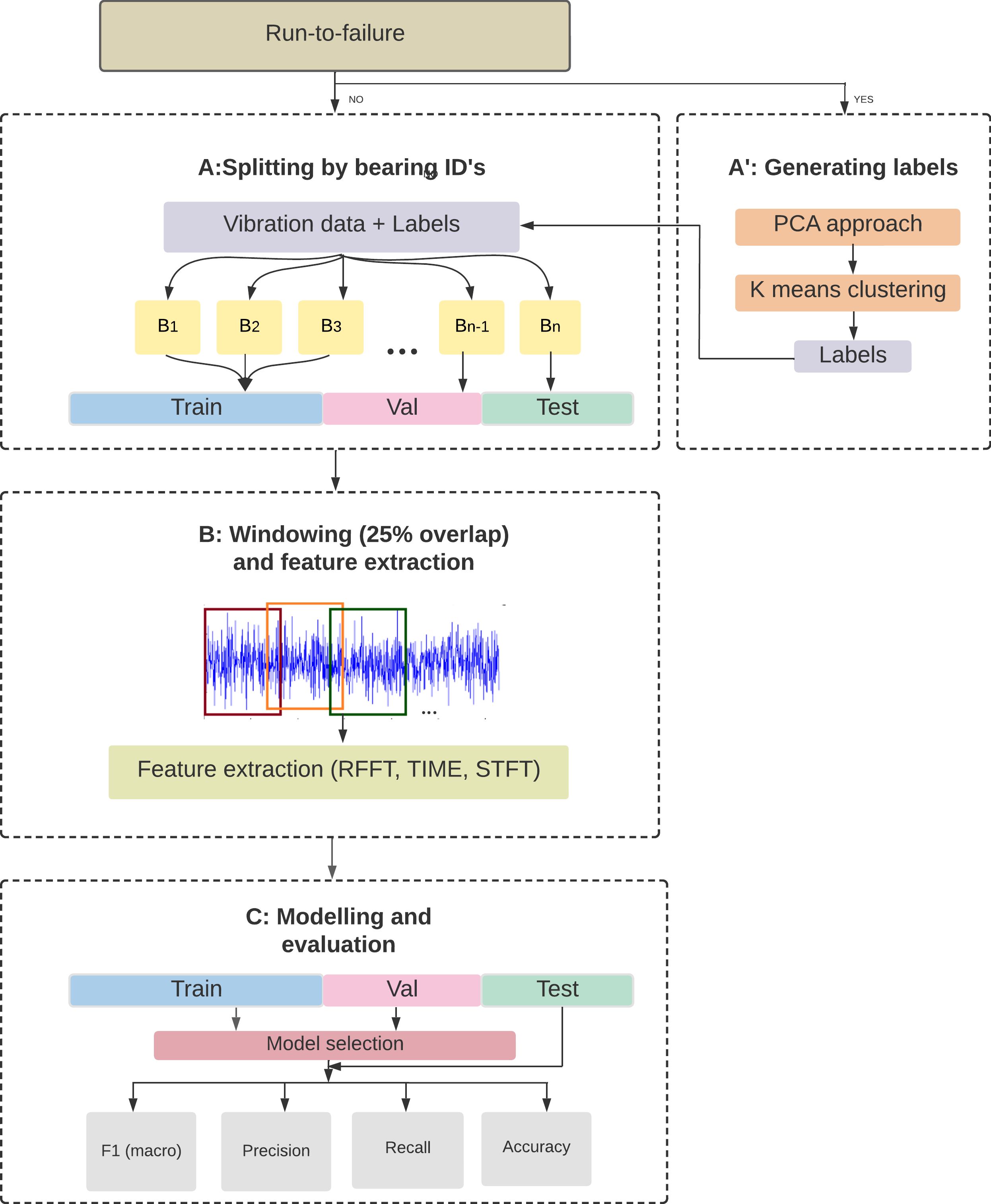}
\caption{Workflow of approach. Our model development follows a three step process denoted by boxes A, B, and C. The first step, box A starts with creating data partitions for training, validation and testing, however run-to-failure experiments requires an additional label generation process preceding this step as shown in box A'. The second step, box B, represents data pre-processing and preparation. In this step after windowing the data appropriately, several features are extracted. The final step, box C represents model fitting, and performance assessment.}
\label{fig:flowchart}
\end{figure}

Classification techniques were used to determine if an operation is normal or faulty with datasets obtained from accelerated run-to-failure experiments, and induced failures. We consider a binary classification approach to separate faults and normal operation. In datasets where the types of failures are identified (XJTU and CWRU), we use a multi-class classification approach where each fault is considered as a class along with normal operation. The multi-class labels of these two dataset are however different as described in Section \ref{sec:datasets}, and the total number of samples in each fault category after windowing is shown in Table \ref{class_splits}. Our approach for obtaining the labels, splitting the dataset, pre-processing, model development, and performance evaluation is illustrated with a schematic in Figure \ref{fig:flowchart}.

\vspace{-10pt}  

\subsection{Dataset splits}

We split the data collected from each bearing into train (Train), validation (Val) or test (Test) splits as shown in Figure \ref{fig:flowchart} (Block A). We use this split strategy of assigning data from each bearing only to a single partition as it would help us understand the ability of models to generalize to new bearings, a condition necessary for it to be useful in the real world. We also experiment with a random split approach to understand the impact of splits on performance.  Table \ref{table:dataset_splits} shows the IDs of the bearing (as represented in the original study) and how they are assigned to training, validation or test splits for the FEMTO, XJTU, and CWRU datasets. 

\begin{table}[!t]
\caption{Assignment of bearings to Train, Val or Test splits (bearing ID as represented in the original study)}
\begin{center}
\begin{tabular}{p{1cm}|p{2cm}|p{2cm}|p{2cm}}
\hline \hline
\textbf{Dataset} & \textbf{Train Bearing IDs} & \textbf{Val Bearing IDs} & \textbf{Test Bearing IDs} \\ \hline \hline
FEMTO & 1\_1, 1\_2, 2\_1, 2\_2,
3\_1, 3\_2, 1\_3,
2\_3 & 1\_4, 1\_5, 2\_4, 2\_5 & 1\_6, 1\_7, 
 3\_3, 2\_6, 2\_7 \\ \hline
XJTU  & 1\_1, 1\_4, 1\_5,
2\_1, 2\_2, 2\_4, 3\_1, 3\_2, 3\_5 & 1\_3, 2\_5, 3\_3 & 1\_2, 3\_4, 2\_3 \\ \hline
CWRU &12k\_Drive/B, 12k\_Drive/IR, 12k\_Drive/OR, 48k\_Drive/B, 48k\_Drive/IR, 48k\_Drive/OR, Normal
& 12k\_Fan/B/-(007,014), 12k\_Fan/IR/-(007,014), 12k\_Fan/OR/-(007,014), Normal & 12k\_Fan/OR/-(021), 12k\_Fan/B/-(021), 12k\_Fan/IR/-(021), Normal\\ \hline
\end{tabular}
\label{table:dataset_splits}
\end{center}
\end{table}

\subsection{Generating labels for run-to-failure experiments}
\label{subsec:labels}
 We consider two approaches to partition the datasets obtained from run-to-failure experiments into failure(s) and non-failure classes. We consider an approach where the time points after the first instant the accelerometer reading exceeds either 5g or 10g as failure. This naive threshold approach is motivated by how accelerated run-to-failure experiments are terminated at a certain point in time when vibration exceeds a certain level.  We also consider another approach used as a baseline in \shortcite{juodelyte2022predicting} that uses PCA of the spectral data and k-means to generate bearing failure classes as shown in Figure \ref{fig:flowchart} (Block $A^`$). This approach projects the features of the signal into a lower dimensional space using PCA, followed by clustering that data using a k-means algorithm \shortcite{juodelyte2022predicting}. We use this approach with 4 classes for k-means, and  the classes enriched for data points that occur later in time (when the experiments are terminated with a failed bearing) are termed as failure class(es). The rest of the points are considered normal (non-failure).

\subsection{Windowing and feature extraction}

The duration and the number of the vibration waveforms recorded across datasets are different (0.1 secs in FEMTO to upto 10 secs in CWRU for a single waveform). 
We segment a single waveform into windows of 1024 or 2048 points in length as used by \shortcite{peng2021multi}, and with a 25\% overlap between segments as shown in Figure \ref{fig:flowchart} (Block B). Our experimental results however were not sensitive to this choice of window length, and we report performance of classifiers with 2048 points.  After splitting the signals into equal segments, different time domain features (TIME): mean, absolute median, standard deviation, skewness, kurtosis, crest factor, energy, RMS, number of peaks, number of zero crossings, Shapiro test, KL divergence \shortcite{juodelyte2022predicting}; frequency domain frequencies: Real-Valued Fast Fourier Transform (RFFT) \shortcite{juodelyte2022predicting},  and time-frequency: Short Time Fourier Transform (STFT) \shortcite{wang2017virtualization} features are extracted.  We show the total number of feature records/samples available for each split in Table \ref{segment_length}. The proportion of samples available for each split is not consistent across datasets as our splits are based on the bearings, and the number of observations available for each bearing may vary due to differences in either time to failure or duration of the recording. 

\begin{table}[!t]   
	\begin{center}  
	\caption{Number of data samples in Train, Val and Test sets}
	\label{segment_length}
	\begin{tabular}{p{1cm}|p{1.3cm}|p{1.3cm}|p{1.3cm}}
		\hline \hline
		\textbf{Splits} & \textbf{FEMTO} & \textbf{CWRU} & \textbf{XJTU}\\ 
		\hline \hline
		Train &  102212 &	3660 &	98224 \\ 
		\hline
		Val	& 13010 & 555 &	13888 \\ 
		\hline
		Test	& 14748 &210	& 35344  \\ 
		\hline	
	\end{tabular}
	\end{center}
\end{table}

\begin{table*}[!t] 
	\begin{center}  
	\caption{Data samples distribution across different classes in different datasets}
	\label{class_splits}
	\begin{tabular}{c|c|c|c||c|c|c|c}
		\hline \hline
		\multirow{2}{*}{\textbf{Class}} &  \multicolumn{3}{c||}{\textbf{Binary}} & \multicolumn{4}{c}{\textbf{Multiclass}}\\
		\cline{2-8}
		& \textbf{FEMTO} & \textbf{XJTU} & \textbf{CWRU} & \multicolumn{2}{c|}{\textbf{XJTU}}& \multicolumn{2}{c}{\textbf{CWRU}} \\ 
		\hline \hline
\centering &&&&OR &4928 &OR&2148\\
\cline{5-8}
&&&&IR&1152&IR&1016\\
\cline{5-8}
Failure&18972 &12032 & 4215&CAGE&3136&BALL&1051\\
\cline{5-8}
&&&&IR and OR&32&&\\
\cline{5-6}
&&&&COBI&2832&&\\
\hline
Normal&110998 &135424 & 210 &Normal & 135376 &Normal & 210\\
\hline
\end{tabular}
\end{center}
\end{table*}

\subsection{Modeling and evaluation}
We train SVM, RF, Logistic Regression (LR), Naive Bayes (NB), and Multi-Layer Perceptron (MLP) models with RFFT, TIME and STFT features. We attempted hyper-parameter tuning of SVM, LR, NB, and MLP, however after our experiments, we noticed the default parameter settings of SVM (\textit{C=1.0, kernel='rbf', class\textunderscore weight='balanced'}), LR (\textit{C=1.0, class\textunderscore weight='balanced'}), NB (\textit{priors=None}) in sklearn\footnote{https://scikit-learn.org/stable/} provided the best performance. We train the MLP model with a three fully connected and batch normalization layers along with ReLU activation and softmax cross-entropy loss using PyTorch\footnote{https://pytorch.org/}. We also use a baseline dummy classifier from sklearn (denoted DUMMY) with a 'stratified' parameter that only considers the class probability distribution without any dependence on the input features.

We finally evaluate the fault classification models using the well established metrics in ML literature (Accuracy ($Acc$), Precision ($Prec$), Recall ($Rec$), F-score ($F$), and F1 macro ($F_{mac}$)) for binary and multi-class classification \shortcite{grandini2020metrics} as shown in Figure \ref{fig:flowchart} (Block C).

\begin{figure}[h]
\centering
\includegraphics[width=\columnwidth]{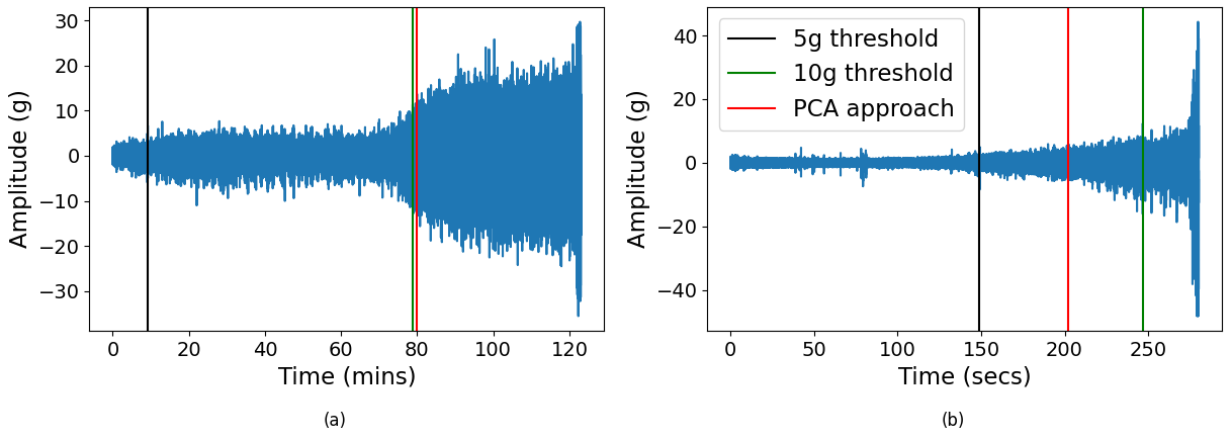}
\caption{Label generation using the threshold and PCA approaches in the run-to-failure datasets (a) XJTU (b) FEMTO.}
\label{fig:failure}
\end{figure}

\section{Results}
In this section we investigate the approaches to identify failures in run-to-failure experiments, performance of binary and multi-class fault classifiers, and the influence of the splits, choices that affect the development of fault classification models.
\label{results}

\subsection{Choice of identifying the failure and non-failure in run-to-failure experiments}
We generated the labels for run-to-failure data using a naive threshold and PCA approaches in Figure \ref{fig:failure}. This figure shows the vibration signal amplitude over time for the XJTU and FEMTO run-to-failure datasets. One can observe this threshold of first occurrence of either 5g (black line) or 10g (green line) seems to work reasonably in XJTU dataset (see Fig 2a), separating the regions where the amplitude ranges are rising, from the one where it is roughly constant. In fact, it works well across most bearings in this dataset (results not shown). However this naive approach doesn't work well on FEMTO dataset where 5g seems better than 10g on separating the signal where vibration amplitudes starts to increase (see Fig 2b), and doesn't remain consistent across other bearings. The amplitude ranges of the vibration across datasets also changes (upto 40 g in FEMTO versus 30 g in XJTU), which also depends on the load and other operating conditions making a threshold based approach not consistent.  The approach of using PCA followed by k-means clustering (red line) seems to separate the two regions of low amplitude and high amplitude signals better. The results of training various multi-class classification models on the data partitioned using the 10g threshold, and the PCA approach in XJTU dataset is shown in Tables \ref{10g} and \ref{XJTU} respectively. These results though not comparable, as the test sets are different, show similar performance. We use the PCA approach that is generalizable across datasets, and illustrate the results in the remaining sections of the paper.

\begin{table}[h] 
	\begin{center}  
	\caption{Multiclass classification results on XJTU dataset using 10g threshold}
	\label{10g}
	\begin{tabular}{ c|c|c|c|c|c}
		\hline \hline
		\multirow{2}{*}{\textbf{Models}} & \multirow{2}{*}{\textbf{Features}} &  \multicolumn{4}{c}{\textbf{Multiclass}}\\
		\cline{3-6}
		&& \textbf{$Acc$} & \textbf{$F_{mac}$} & \textbf{$Prec$} & \textbf{$Rec$} \\ 
		\hline \hline
\multirow{3}{*}{SVM}& RFFT & 0.757   &0.272   &0.257   &0.362  \\
\cline{2-6}
&TIME& 0.869   &0.250  & \textbf{0.385}  & 0.326 \\
\cline{2-6}
&STFT &  0.336&   0.191 &  0.255 &  0.258\\
\hline
\multirow{3}{*}{NB} & RFFT&  0.849&   0.273   &0.249   &0.326  \\
\cline{2-6}
&TIME  &0.869   &\textbf{0.295}   &0.265   &0.354 \\
\cline{2-6}
& STFT &  0.852 &  0.276 &  0.252  & 0.327 \\
\hline
\multirow{3}{*}{RF} &RFFT &  \textbf{0.874} &  0.286&   0.257&   \textbf{0.387 } \\
\cline{2-6}
& TIME  &  0.841   &0.282   &0.257   &0.383\\
\cline{2-6}
& STFT &  0.837 &  0.282 &  0.257 &  0.379 \\
\hline
\multirow{3}{*}{MLP} & RFFT& 0.869 &  0.245 &  0.221 &  0.293\\ 
\cline{2-6}
&TIME& 0.839 &  0.235 &  0.220 &   0.257\\
\cline{2-6}
& STFT& 0.856 &  0.239 &   0.216 &  0.242 \\ 
\hline
\end{tabular}
\end{center}
\end{table}

\begin{table}[!th]  
	\begin{center}  
	\caption{Binary classification results on FEMTO dataset}
	\label{FEMTO-binary}
	\begin{tabular}{ c | c | c | c | c | c }
		\hline \hline
		\multirow{2}{*}{\textbf{Models}} & \multirow{2}{*}{\textbf{Features}} & \multicolumn{4}{c}{\textbf{Binary}}\\
		\cline{3-6}
		&& \textbf{$Acc$} &  \textbf{$F$} &\textbf{$Prec$} & \textbf{$Rec$}\\ 
		\hline \hline
\multirow{3}{*}{DUMMY}&RFFT& 0.878&0.492&0.499&0.498\\
\cline{2-6} 
&TIME& 0.873&0.491&0.499&0.497\\
\cline{2-6} 
&STFT & 0.879&0.496&0.502&0.505\\
\hline
\multirow{3}{*}{SVM} & RFFT    &0.950 &  0.732 &  0.931 &  0.666 \\
\cline{2-6} 
&  TIME   &   0.946 &  0.684 &  \textbf{0.969} &  0.624    \\
\cline{2-6} 
& STFT &   0.949  & 0.712 &  \textbf{0.969}&  0.646 \\
\hline
\multirow{3}{*}{NB} &RFFT&  0.987 &  0.954 &  0.933 &  \textbf{0.977}    \\
\cline{2-6} 
& TIME   &  0.938&  0.641 &  0.860 &  0.596 \\
\cline{2-6} 
& STFT &   \textbf{0.991}  & \textbf{0.966} &  0.956  & \textbf{0.977} \\
\hline
\multirow{3}{*}{RF}&  RFFT &   0.936  & 0.585  & 0.960   &0.557  \\ 
\cline{2-6} 
& TIME &0.941  & 0.643 & 0.965  & 0.594\\
\cline{2-6} 
& STFT &   0.938 & 0.605 &  0.962 &  0.569 \\
\hline
\multirow{3}{*}{MLP}& RFFT& 0.938  & 0.618 &  0.894 &  0.579\\
\cline{2-6} 
&TIME& 0.944 &  0.664 &  0.967 & 0.609\\
\cline{2-6} 
&STFT & 0.946 &  0.683 &  0.968 &   0.623\\
\hline
	\end{tabular}
	\end{center}
\end{table}

\begin{table*}[h] 
	\begin{center}  
	\caption{Binary and multiclass classification results on XJTU dataset}
	\label{XJTU}
	\begin{tabular}{ c|c|c|c|c|c||c|c|c|c }
		\hline \hline
		\multirow{2}{*}{\textbf{Models}} & \multirow{2}{*}{\textbf{Features}} &  \multicolumn{4}{c||}{\textbf{Binary}} & \multicolumn{4}{c}{\textbf{Multiclass}}\\
		\cline{3-10}
		&& \textbf{$Acc$} & \textbf{$F$} & \textbf{$Prec$} & \textbf{$Rec$}& \textbf{$Acc$} & \textbf{$F_{mac}$} & \textbf{$Prec$} & \textbf{$Rec$} \\ 
		\hline \hline
\multirow{3}{*}{DUMMY}& RFFT & 0.832 &   0.495 &  0.499 &   0.499  &0.823  & 0.158  & 0.162 &  0.160 \\
\cline{2-10}
&TIME&0.833  &  0.496 &   0.500  & 0.500  & 0.827&0.162 &  0.170  & 0.164 \\
\cline{2-10}
&STFT &  0.831 &   0.492 &   0.495 &   0.497&   0.827 &  0.162 &  0.169 &  0.163 \\
\hline
\multirow{3}{*}{SVM}& RFFT & 0.968 &   \textbf{0.930} &  0.893 &   \textbf{0.979}  &0.892  & 0.266  & 0.242 &  0.351 \\
\cline{2-10}
&TIME&0.944 &   0.846  &  0.897 &   0.810  & 0.885  & 0.234 &  0.220  & 0.254 \\
\cline{2-10}
&STFT &  \textbf{0.969} &   0.925 &   0.924 &   0.926&   0.889 &  0.308 &  0.286 &  \textbf{0.353} \\
\hline
\multirow{3}{*}{NB} & RFFT&  0.950  &  0.896 &   0.848  &  0.971 & 0.728  & 0.251  & 0.243 &  0.297 \\
\cline{2-10}
&TIME  & 0.954 &   0.903 &   0.857 &   0.973  & 0.823  & 0.195  & 0.192 &  0.222 \\
\cline{2-10}
& STFT &   0.952 &   0.900 &   0.854 &   0.973&   0.740 &  0.258  & 0.249 &  0.299 \\
\hline
\multirow{3}{*}{RF} &RFFT & 0.912 &   0.674 &   0.931 &   0.624 &   0.884 &  0.235  & \textbf{0.471}  & 0.250  \\
\cline{2-10}
& TIME  &    0.963  &  0.921  &  0.880  &  \textbf{0.979}   &0.882 &  0.196  & 0.193  & 0.205\\
\cline{2-10}
& STFT &    0.907 &   0.644  &  \textbf{0.935} &   0.602 &  0.888 &  0.293 &  0.468 &  0.283 \\
\hline
\multirow{3}{*}{MLP} & RFFT& 0.943  & 0.866 &  0.853 &  0.880 & \textbf{0.902} &  \textbf{0.311} &  0.294 &  0.332\\ 
\cline{2-10}
&TIME& 0.915  & 0.701&   0.915&   0.656& 0.883 &  0.253 &  0.245 &   0.263\\
\cline{2-10}
& STFT&0.947  & 0.880 &  0.856&   0.910&  0.884 &  0.264 &   0.258 &  0.269 \\ 
\hline
\end{tabular}
\end{center}
\end{table*}

\begin{table*}[!t] 
	\begin{center}  
	\caption{Binary and multiclass classification results on CWRU dataset}
	\label{CWRU}
	\begin{tabular}{ c|c|c|c|c|c||c|c|c|c }
		\hline \hline
		\multirow{2}{*}{\textbf{Models}} & \multirow{2}{*}{\textbf{Features}} &  \multicolumn{4}{c||}{\textbf{Binary}} & \multicolumn{4}{c}{\textbf{Multiclass}}\\
		\cline{3-10}
		&& \textbf{$Acc$} & \textbf{$F$} & \textbf{$Prec$} & \textbf{$Rec$}& \textbf{$Acc$} & \textbf{$F_{mac}$} & \textbf{$Prec$} & \textbf{$Rec$} \\ 
		\hline \hline
\multirow{3}{*}{DUMMY}& RFFT &0.843 &  0.486 &  0.529 &  0.506  &0.252 &  0.208 &  0.437 &  0.225  \\
\cline{2-10}
&  TIME  &  0.843 &  0.486 &  0.529 &  0.506  &  0.262  & 0.223 &  0.232 &  0.237    \\
\cline{2-10}
& STFT &   0.838 &  0.508 &  0.556 &  0.517 &   0.262 &  0.207 &  0.240 &  0.233 \\
\hline
\multirow{3}{*}{SVM}& RFFT &0.976 &  0.954 &  0.929 &  0.986  &0.657 &  0.628 &  0.614 &  0.700  \\
\cline{2-10}
& TIME  &  0.857  & 0.462 &  0.429  & 0.500 &  0.495  & 0.315 &  0.251 &  0.433    \\
\cline{2-10}
& STFT &   0.981 &  0.963 &  0.941 &  0.989 &   \textbf{0.700} &  0.658 &  0.622 &  \textbf{0.738} \\
\hline
\multirow{3}{*}{NB}& RFFT & \textbf{1.000}  & \textbf{1.000}  & \textbf{1.000} &  \textbf{1.000} &   0.695  & \textbf{0.693} &  \textbf{0.808}  & 0.733    \\
\cline{2-10}
& TIME &  0.295 &  0.233 &  0.343 &  0.186    &  0.067  & 0.043 &  0.034 &  0.058 \\
\cline{2-10}
& STFT &   \textbf{1.000} &  \textbf{1.000} &  \textbf{1.000} &  \textbf{1.000} &   0.467 &  0.518 &  0.521 &  0.533 \\
\hline
\multirow{3}{*}{RF}& RFFT &   0.857 &  0.462 &  0.429 &  0.500 &   0.457 &  0.296 &  0.236 &  0.400   \\
\cline{2-10}
& TIME &0.857 &  0.462 &  0.429  & 0.500    &0.267 &  0.177 &  0.142 &  0.233\\
\cline{2-10}
& STFT &   0.976 &  0.948 &  0.986 &  0.917&   0.338 &  0.190 &  0.231 &  0.296 \\
\hline
\multirow{3}{*}{MLP}&  RFFT &  0.924  & 0.871 &  0.826  & 0.956  &  0.471  & 0.441 &  0.416 &  0.537 \\
\cline{2-10}
& TIME  &   0.857 &  0.462 &  0.429 &  0.500  &   0.362 &  0.229 &  0.186  & 0.317 \\
\cline{2-10}
& STFT &  0.857  & 0.462  & 0.429 &  0.500 &  0.624  & 0.575 &  0.510 & 0.671\\
\hline
	\end{tabular}
	\end{center}
\end{table*}

\begin{table*}[h]   
	\begin{center}  
	\caption{Multiclass classification results on XJTU dataset with bearing split and random split}
	\label{XJTU_splits}
	\begin{tabular}{ c|c|c|c|c|c||c|c|c|c }
		\hline \hline
		\multirow{2}{*}{\textbf{Model}} & \multirow{2}{*}{\textbf{Feature}}&\multicolumn{4}{c||}{\textbf{Bearing split}} & \multicolumn{4}{c}{\textbf{Random split}}\\
		\cline{3-10}
		&& \textbf{$Acc$} & \textbf{$F_{mac}$} & \textbf{$Prec$} & \textbf{$Rec$}& \textbf{$Acc$} & \textbf{$F_{mac}$} & \textbf{$Prec$} & \textbf{$Rec$} \\ 
		\hline \hline
		 SVM & STFT &  0.890 &  0.309 &  0.286 & 0.355 & 0.988 &  0.779  & 0.787  & 0.774 \\ 
		\hline
		NB	& RFFT & 0.728  & 0.251  & 0.243 &  0.297 &0.910 &  0.616 &  0.515 &   0.899\\ 
		\hline	
	\end{tabular}
	\end{center}
\end{table*}

\begin{table*}[h]   
	\begin{center}  
	\caption{Multiclass classification results on CWRU dataset with bearing split and random split}
	\label{CWRU_splits}
	\begin{tabular}{ c|c|c|c|c|c||c|c|c|c }
		\hline \hline
		\multirow{2}{*}{\textbf{Model}} & \multirow{2}{*}{\textbf{Feature}}&\multicolumn{4}{c||}{\textbf{Bearing split}} & \multicolumn{4}{c}{\textbf{Random split}}\\
		\cline{3-10}
		&& \textbf{$Acc$} & \textbf{$F_{mac}$} & \textbf{$Prec$} & \textbf{$Rec$}& \textbf{$Acc$} & \textbf{$F_{mac}$} & \textbf{$Prec$} & \textbf{$Rec$} \\ 
		\hline \hline
		 SVM & STFT & 0.700 &  0.658 &  0.622 &  0.738& 0.851  & 0.865 &  0.925 &  0.834  \\ 
		\hline
		NB	& RFFT & 0.695  & 0.693 &  0.808  & 0.733  & 0.858  &   0.833 &  0.902 &  0.792 \\ 
		\hline	
	\end{tabular}
	\end{center}
\end{table*}

\subsection{Binary and multi-class fault classifiers}


We show the results of binary and multi-class setups on FEMTO, XJTU, and CWRU datasets in Tables \ref{FEMTO-binary}, \ref{XJTU}, and \ref{CWRU} respectively. We observe the performance of the binary classification is in general better across the methods than that of the multi-class classifier. This likely arises due to the vibration signatures of non-failure classes being distinct from failures, but differences in signatures between failure classes might be more subtle. In the binary setup, we observe from Tables \ref{FEMTO-binary}, \ref{XJTU}, \ref{CWRU}, that NB with STFT/RFFT features performs best on FEMTO and CWRU datasets, and SVM with STFT/RFFT features performs best on XJTU with an $F$ score roughly 43\% greater than the dummy classifier.

The number of samples available for both the binary and multi-class classification is highly imbalanced as shown Table \ref{class_splits}, however the metric used to assess models widely in the literature is just accuracy \shortcite{zhao2020deep,zhao2021improved, schwendemann2021survey, zhao2020fault, neupane2020bearing}. Clearly our results both in the binary and multi-class formulations show even though the accuracy metrics are high, other metrics are relatively lower. Precision and recall metrics are useful in imbalanced binary problems, however the average of this still might not be enough with imbalanced  multi-class problems, and we use $F_{mac}$ score for interpreting these problems. Further to understand where we stand with respect to the performance of a given classifier we compare it to the results of the dummy classifier that classifies a sample randomly to a category with a probability proportional to the number of samples in that category.

In the multi-class classification setup, we observe from Table \ref{XJTU}, MLP model with RFFT features performs well on XJTU with roughly 15\% $F_{mac}$ more than the dummy classifier. We also observe SVM and RF with STFT trails the best method by just 0.3\% and 1.8\% respectively. On CWRU dataset, NB with RFFT performed well as shown in the Table \ref{CWRU} which is 48\% more than the dummy classifier. This improvement over the baseline could possibly be due to the stark differences in vibration data observed due to the injection of faults in novel bearings as opposed to the continuous progression towards failure in run-to-failure experiment like XJTU.

\subsection{Importance of bearing information while doing splits}

We observe our performances are conservative as compared to various studies that reported multi-class classification performance on these datasets \shortcite{zhao2020deep,zhao2021improved, schwendemann2021survey} which used a random split strategy of assigning feature records from the same bearing to more than one of Train, Val and Test splits, (e.g. features from bearing 2\_1 in XJTU could be part of Train, Val and Test splits). In order to understand how well the model trained on data from a few given bearings will perform in the `field` (i.e how well will the model perform on vibration data from a new bearing under similar but not identical settings), we will have to tailor our model assessment strategy to reflect this setting.  Our split strategy of assigning the bearings to just one split more closely resembles this scenario of assessing the classifier performance in the real world. We however generated results on the random split to show we are able to generate performance similar to what has been reported in the literature. Although the test datasets with the bearing split and random split are not comparable as the test partitions are different. We observe the performances of SVM with STFT and NB with RFFT on the bearing split are conservative by 47.0\% and 36.5\% $F_{mac}$  on XJTU datasets (Table \ref{XJTU_splits}). We observe a similar trend on the CWRU where SVM with STFT and NB with RFFT on the bearing split are conservative by 20.7\% and 14.0\% $F_{mac}$ (Table \ref{CWRU_splits}). We have also looked at class confusion matrices to understand which classes perform better, and observed that normal category does better than other specific faults. We hypothesize it is likely due to the number of samples available in each category to train the model with respect to the variability within a given class.

\section{Conclusion}
\label{sec:conclusion}

In our work we investigated the impact of various choices that are important to reliably develop models, and deploy them in practical settings using publicly-available vibration datasets. Our problem formulation of fault classification into binary or multi-class demonstrates that performances are higher in the binary setting as compared to the multi-class setting. We also observe injected failures are easy to separate over segmented run-to-failure datasets. We also show vibration datasets are heterogenous in the amount of data available for different failures, and its important to look at metrics that handles the multi-class scenario and imbalance in the dataset. Our investigation also showed the robustness of the PCA based approach as compared to simple threshold in handling multiple datasets. We also demonstrate the importance of splitting by bearing as opposed to reporting it on random splits. We also show the usefulness of baseline dummy classifiers that uses class statistics, well known strategy used in practical ML settings, for developing bearing fault classification approaches, and recommend other research studies report their performance on this baseline along with other baselines.  

This work helped underscore the importance of several key choices to reliably develop models, and deploy them in practical settings for PHM using the acceleration data. It is important for future studies to investigate other choices not considered in this study such as different features obtained using other transformations, and signal representations learned from data using more advanced embedding approaches. Our study considered only a baseline MLP classifier, and we leave other modern neural architectures like convolutional, recurrent networks with various learning paradigms for future work.

\vspace{-10pt}

\bibliographystyle{apacite}
\bibliography{references}

\end{document}